%% file: main.tex
\documentclass[10pt,twocolumn,letterpaper]{article}

\usepackage{cvpr}              % To produce the CAMERA-READY version
\usepackage[table]{xcolor}
\usepackage{multirow} 

\usepackage{pifont}

\newcommand{\cmark}{\ding{51}}  % ✓
\newcommand{\xmark}{\ding{55}}  % ✗

\input{preamble}

\definecolor{cvprblue}{rgb}{0.21,0.49,0.74}
\usepackage[pagebackref,breaklinks,colorlinks,allcolors=cvprblue]{hyperref}

\def\paperID{***} % *** Enter the Paper ID here
\def\confName{CVPR}
\def\confYear{2026}

\title{MuKV: Multi-Grained KV Cache Compression for Long\\ Streaming Video Question-Answering}

\author{
Junbin Xiao\textsuperscript{1,2}\thanks{Equal Contribution.},
\quad Jiajun Chen\textsuperscript{2*},
\quad Tianxiang Sun\textsuperscript{2},  
\quad Xun Yang\textsuperscript{1}, 
\quad Angela Yao\textsuperscript{2} \\
\textsuperscript{1} University of Science and Technology of China, \textsuperscript{2} National University of Singapore \\
{\tt\small{junbinxiao@ustc.edu.cn, chen.jiajun@u.nus.edu, angela.yao@nus.edu.sg}}
}

\begin{document}
\maketitle
\input{sec/0_abstract}    
\input{sec/1_intro}

\input{sec/2_related_work}
\input{sec/3_method}

\input{sec/4_experiment}
\input{sec/5_conclusion}
{
    \small
    \bibliographystyle{ieeenat_fullname}
    \bibliography{main}
}
\input{sec/X_suppl}

\end{document}

%% file: preamble.tex
\usepackage{soul}

%% file: sec/0_abstract.tex
\begin{abstract}
Long streaming video QA remains challenging due to growing visual tokens and limited reasoning length of large language models (LLMs). KV-caching stores the Key-Value (KV) of the historical tokens via LLM prefill and enables more efficient streaming QA.  However, existing methods cache every one or two frames, causing redundant memory usage and losing fine-grained spatial details within frame or temporal contexts across frames. This paper proposes MuKV, a method that features a multi-grained KV cache compression module and a semi-hierarchical retrieval approach to improve both efficiency and accuracy for long streaming VideoQA. For the offline KV cache, MuKV extracts visual representations at patch-, frame-, and segment-levels. The multiple levels of granularity preserve both local cues and global temporal context, while maintaining efficiency with a dual signal token compression mechanism guided by self-attention and frequency. For online QA, MuKV designs a semi-hierarchical retrieval method to retrieve relevant KV caches for answer generation. Experiments on long-streaming VideoQA benchmarks show that MuKV significantly improves answer accuracy, without sacrificing memory and online QA efficiency. Moreover, our compression mechanism alone brings consistent benefits across answer accuracy, memory, and QA efficiency over baselines, showcasing highly effective contribution. 
\end{abstract}

%% file: sec/1_intro.tex
\section{Introduction}
Multimodal large language models (MLLMs) have enabled remarkable progress in VideoQA \cite{alayrac2022flamingo,maaz2023video,zhang2023video,lin2023video,li2024mvbench,wang2025videochat,zhang2025videollama,bai2025qwen3,wang2025internvl3}. Yet, most of the advances are made on understanding relatively short or offline videos with determinate lengths~\cite{xu2017video,yu2019activitynet,xiao2021next,fu2025video}. Extending such capabilities to online streaming videos of unconstrained length remains a significant challenge. The core difficulty lies in the linear growth of visual tokens with time, which quickly exceeds the reasoning context window of LLMs, and leads to inefficiency and reduced answer accuracy.

\begin{figure}
  \centering
  \begin{subfigure}{0.48\linewidth}
    \includegraphics[width=\textwidth]{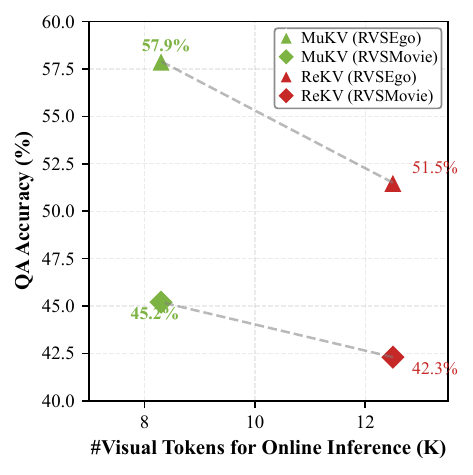}
    \caption{QA efficiency and accuracy.}
    \label{fig:short-a}
  \end{subfigure}
  \hfill
  \begin{subfigure}{0.48\linewidth}
    \includegraphics[width=\textwidth]{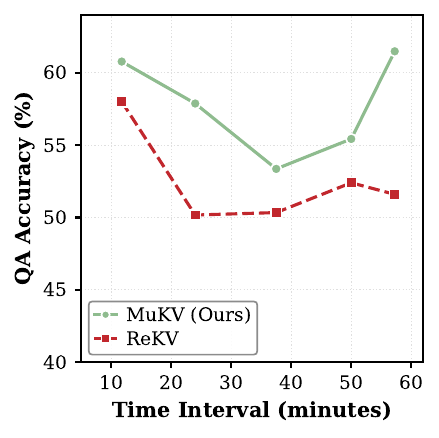}
    \caption{Accuracy over time.}
    \label{fig:short-b}
  \end{subfigure}
  \caption{Comparison between MuKV and previous arts. (a) MuKV improves ReKV's answer accuracy, without increasing online QA time and offline KV storage. (b) The advantage gets strengthened over time in streaming video QA. }
  \label{fig:intro}
  \vspace{-3mm}
\end{figure}

\begin{figure*}[t!]
\centering
\includegraphics[width=1.0\textwidth]{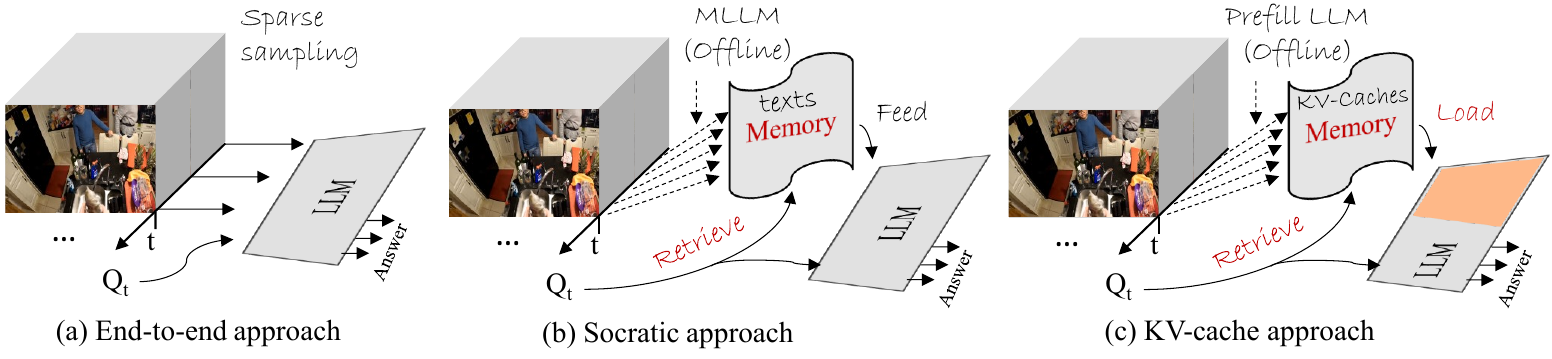}
\vspace{-6mm}
\caption{Illustration of different approaches for streaming video QA. (a) The end-to-end approach trades off visual details for long-ranged modeling. (b) The Socratic approach suffers from online efficiency since it consumes full LLM computation online. (c) The KV-cache approach enables more efficient online QA via exemption of online KV-computation of past visual tokens. }
\label{fig:related}
\vspace{-3mm}
\end{figure*}

To address this challenge, primary approaches focus on extending LLM context window \cite{zhang2024long,weng2024longvlm,xu2024pllava,shen2024longvu,wang2025internvideo2,yang2025qwen3}, \eg, via token compression, though this may require large-scale pretraining, incur quadratic computation costs, and trade off visual details for long-ranged modeling. As such, this strategy is limited in its applicability for streaming QA. A second approach uses decoupled \textit{offline memory} and \textit{online retrieval} mechanisms \cite{fan2024videoagent,wang2024videoagent,qian2024streaming,chatterjee2025memory,qin2025question,distreaming,kim2025infinipot}, with the LLM context window unaltered. They offline store either visual descriptions \cite{zhang2024simple,wang2025videotree,qin2025question,xiao2025unleashing,chatterjee2025memory}, embeddings \cite{chatterjee2025memory,zhang2025flash}, or Key-Value (KV) caches \cite{distreaming,kim2025infinipot} of the past video streams, and online retrieve a subset of relevant information for answering when a question triggered. Among them, KV-caching enables training-free and efficient online answer decoding (without redundant KV computation of history tokens), showing promising streaming QA applicability.

However, existing KV-cache approaches \cite{distreaming,kim2025infinipot} are limited to per-frame caching, the frame-level representations alone may hardly encode region-level visual details and cross-frame temporal contexts. Also, the linearly increased KV cache brings significant redundancy for storage, which may affect retrieval and consequently QA performance, leaving open challenges for effective KV-cache memory, compression, and retrieval \cite{kim2025infinipot,xu2025streamingvlm,ning2025livevlm,chen2025streamingtom}.

In this paper, to improve KV-cache memory and retrieval, we propose MuKV, a novel method that underscores a multi-grained KV-cache compression module and a semi-hierarchical retrieval mechanism for more effective long streaming VideoQA. For offline memory, MuKV extracts video KV caches at multiple granularity levels: patch, frame, and segment — to preserve spatial details and maintain temporal contexts. Importantly, to reduce the cache for efficient memory and retrieval, MuKV proposes a Dual signal KV-cache Compression (DCP) mechanism guided by self-attention importance and token frequency to prune redundant KV caches. For online QA, MuKV designs a semi-hierarchical retrieval strategy that first retrieves in parallel among KV caches across all three granularities with the question as query, and then reranks the top-ranked KV tokens via coarse-to-fine cross-granularity retrieval. The final top-ranked KV caches are loaded into LLMs along with the question for answer generation.
Extensive experiments demonstrate that MuKV achieves significant accuracy improvements in QA over long video streams, without sacrificing memory and online QA efficiency (\cref{fig:intro}). 

To summarize our contributions:
\begin{itemize}
    \item We propose MuKV, a streaming VideoQA approach that highlights a novel offline multi-grained KV-cache compression module and an online semi-hierarchical retrieval strategy to improve answer accuracy, yet without sacrificing memory and online QA efficiency.
    \item We propose DCP compression mechanism, and demonstrate that frequency signals along with self-attentions are effective indicators for video KV-cache compression.
    \item Our method shows improved advantage as the video stream continues to grow longer.
\end{itemize}

%% file: sec/2_related_work.tex
\section{Related Work}
\subsection{MLLMs for Long Streaming VideoQA}
Large language models (LLMs) \cite{hurst2024gpt,vicuna2023,dubey2024llama,yang2025qwen3} has stimulated a surge of multimodal LLMs (MLLMs) that enable freely chat with images and videos \cite{xiao2025videoqa,maaz2023video,zhang2023video,lin2023video,li2024llava,wang2025videochat,wang2025internvl3}. 
Despite the success, the end-to-end MLLMs (\cref{fig:related}(a)) often require large-scale pretraining and trade off visual details for long video modeling \cite{weng2024longvlm,zhang2024long,shen2024longvu,zhang2025flash,shu2025video}, with also severely reduced efficiency for the quadratic increase of computation, making them less-practical in coping with dynamic video streams.

In contrast, Socratic or agentic approaches \cite{zengsocratic,fan2024videoagent,min2024morevqa,wang2024videoagent,zhang2024simple} leverage pretrained MLLMs for long streaming VideoQA by introducing offline memory and online retrieval modules (\cref{fig:related}(b)) to store and recall past visual information (\eg, visual descriptions or embeddings). While effective, both descriptions and embeddings demand full online LLM computation, leading to high inference latency. In addition, descriptions alone may not be accurate and sufficient to answer questions in streaming QA practice. The recent ReKV~\cite{distreaming} introduces storing the KV cache as a compromise between visual embeddings and language descriptions (\cref{fig:related}(c)).  However, per-frame KV caches 
are insufficient for video modeling (\eg, lacking spatial details and temporal contexts) whilst consuming large amounts of storage that in turn reduces retrieval and QA performance.
 
Our approach, MuKV, follows the fashion of offline KV-cache memory and online retrieval to achieve streaming QA (\cref{fig:related}(c)). In contrast to ReKV \cite{distreaming}, which indiscriminantly stores per-frame KV caches without considering information granularity and redundancy, 
we introduce to compactly represent the video at multiple granularities on a per-segment basis achieved by a novel KV-cache compression mechanism, 
Such designs allow MuKV to retain better the spatial/temporal fidelity of videos for improved QA, yet achieve sub-linearly increased memory to account for continuous video streamings. 

\subsection{KV-Cache Compression}
To compress the KV-cache, existing algorithms focus on token quantization \cite{hooper2024kvquant, liu2024cachegen} and pruning \cite{xiaoefficient, tang2024razorattention, jo2025fastkv}. 
Quantization may introduce approximation errors and hurt prediction accuracy. Thus, we focus on pruning to reduce redundancy. The popular indicators for token pruning are self-attention score \cite{kim2025infinipot,yang2025streammem, fu2024not,jo2025fastkv} and token similarity \cite{chen2024image, kim2025infinipot}. While self-attention scores are easy to obtain (during LLM prefilling), token similarity often introduces non-trivial computation costs because of pairwise token comparison. 
Recent work InfiniPot \cite{kim2025infinipot} also introduces Value Norm as an effective and more efficient indicator. However, it limits in selecting salient frame regions in spatial domain, failing to analyze across temporal frames and segments.

In MuKV, we introduce an equally efficient but more generalized indicator -- frequency, with the intuition that token frequency distribution within a frame and across frames helps localize semantically important regions and moments respectively. The closest work FreqKV~\cite{kai2025freqkv} also uses frequency as an indictor for KV-cache compression, though they target reducing text tokens of high frequency for online natural language decoding. MuKV differs fundamentally by focusing on video KV cache compression, also at multi-granularity levels of spatial patches, temporal frames, and segments, to reduce offline memory redundancy, in which we find that tokens of high frequency to be more important.

\subsection{KV-Cache Retrieval}
KV-cache retrieval retrieves a subset of token KV values in the cache %that are highly 
relevant to the query for decoding the answer. Existing progresses on pure text QA \cite{wang2025fier,liu2025freekv} directly exploit self-attention scores. For cross-modal retrieval in video QA, pioneer approaches \cite{distreaming,ning2025livevlm,kim2025infinipot} consider cross-modal cosine similarity. However, they apply parallel retrieval among all tokens, 
which may bring noises when copping with multi-grained information. While hierarchical retrieval \cite{wang2025videotree} shows benefits, it suffers from error propagation if the top-level retrieval is wrong. For balance, we design a semi-hierarchical retrieval method by firstly performing hierarchy-agnostic parallel retrieval and then reranking the retrieved visual tokens at lower granularity via cross-grain hierarchical retrieval. Our experiments show that this semi-hierarchical approach effectively reduces noises and improves answer accuracy, despite slightly reducing efficiency due to reranking. 

%% file: sec/3_method.tex
\section{Method}
\begin{figure}[!t]
\centering
\includegraphics[width=0.48\textwidth]{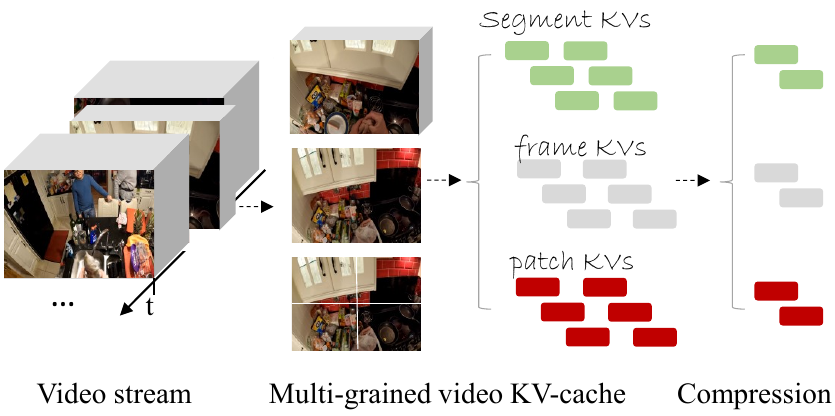}
\caption{Illustration of multi-grained video KV cache compression in offline memory.}
\label{fig:mukv}
\vspace{-3mm}
\end{figure}
\subsection{Problem and Method Overview}
\label{sec:overview}

\paragraph{Problem Formulation.} Given a streaming video $\mathcal{V} = \{f_t\}_{t=1}^{T}$ arriving sequentially and a series of user questions $\mathcal{Q}=\{q_i\}_{i=1}^{M}$ triggered at different timestamps $\{t_i\}_{i=1}^{M}$, streaming VideoQA aims to generate answers $\mathcal{A}=\{a_i\}_{i=1}^{M}$ according to
the video contents up to $t_i$ for each question. Since the user questions are not known in advance, effective streaming video QA system must
: (i) maximally maintain critical past visual information for answering unknown questions, (ii) within affordable memory budgets, and (iii) enable real-time online QA.

\paragraph{Method Overview.}
To achieve streaming VideoQA, we proposes MuKV which follows the KV-cache framework shown in \cref{fig:related}(c). It mainly includes two modules: offline KV-cache memory to memorize KV caches of past video streams and online KV-cache retrieval to retrieve question relevant video KV caches for online answer generation. On top of this framework, we design \textbf{(1) a multi-grain KV cache compression} module that compactly stores the past video KV-caches of multi-granularities offline, and \textbf{(2) a semi-hierarchical KV cache retrieval} mechanism for efficient and accurate online QA. MuKV is training-free, model-agnostic, and can be integrated seamlessly with any existing Video-LLMs in principle. We describe in detail our two innovative designs in the next subsections.

\subsection{Multi-Grained Video KV Cache}
\label{sec:multigran}
While previous methods \cite{distreaming,kim2025infinipot} treat a video stream as a sequence of frames. We encode a video stream $\mathcal{V}^T = \{v_1, v_2, \ldots, v_T\}$ incrementally at a per-segment basis to embrace video information of multiple granularities within each segment. Specifically, as shown in \cref{fig:mukv}, at each time step $t$, we derive three video representations from the current segment $v_t$: the whole segment $v_t$, the middle frame $f_t$ of this segment, and the patches of the middle frame $p_{f_t}$. In practice, since each visual token in existing MLLMs (\eg, LLaVA-OV \cite{li2024llava}) corresponds to an original frame patch, the three-level of representations can be derived by grouping visual tokens at different scales without multiple times of visual encoder inference. For example, assume there are $F$ frames in each segment and $P$ total patches in each frame, and token $x_i$ represents a single patch, if we partition a frame into $S$ super patches ($S<P$), then $v_t=\{x_i\}_{i=1}^O$($O=P \times F$),  $f_t=\{x_i\}_{i=1}^P$, and  $p_{f_t}=\{x_i\}_{i=1}^{\lfloor P/S \rfloor}$ can effectively represent a segment, a middle frame of the segment, and a super patch of the middle frame, respectively.

To obtain the KV caches for the current segment, the above token representations at each granularity level (\eg, $f_t$), along with all the past KV caches at the same granularity level (\eg, $\{(\mathbf{K}^{(\ell)}_{f_{1:t-1}}, \mathbf{V}^{(\ell)}_{f_{1:t-1]}})\}_{\ell=1}^{L}$), are independently fed into LLM to obtain the KV caches at this time step (\eg, $\{(\mathbf{K}^{(\ell)}_{f_t}, \mathbf{V}^{(\ell)}_{f_t})\}_{\ell=1}^{L}$), where $L$ is the number of transformer layers. At the same time, we obtain attention weights $\{\mathbf{A}^{(\ell)}_{f_t}\}_{\ell=1}^{L}$ as indicators for subsequent KV cache compression. Note that each time step performs three LLM prefills to obtain the three granularity KV caches. Moreover, the three prefills are executed in parallel in practice and incur no additional latency.

\subsection{Dual-Signal KV-Cache Compression}
\label{sec:compression}
Multi-grained video KV caches will result in heavy memory redundancy. For compression, we design a \underline{d}ual-signal KV-cache \underline{c}om\underline{p}ression module (DCP) by jointly considering self-attention score and token frequency. Since our compression is independently applied for KV caches of different granularities. We use the frame-level representations as an example to introduce the details below.

% \paragraph
\textbf{Attention-based Indicator.}
Self-attention score serves as an ideal indicator for token selection, since it naturally reflect a token's importance among the token sequence and is readily available along with KV caches without additional computation. In our implementation, the attention scores of the last LLM layer are chosen as effective indicator, as we empirically find that there is no obvious distribution pattern among the attention scores of the intermediate layers. Concretely, the token importance scores within $f_t$ is obtained by aggregating the last-layer attention across multiple heads of all tokens:
\begin{equation}
    \mathbf{I}_{\text{att}} = \frac{1}{H \cdot P} \sum_{h=1}^{H} \sum_{i=1}^{P} \mathbf{A}^{(L)}_{h,i}, \quad   \mathbf{I}_{\text{att}}\in \mathbb{R}^{P\times 1},
\end{equation}
where $H$ denotes the number of self-attention heads.

% \paragraph
\textbf{Frequency-based Indicator.} While attention score reflects a token's semantic importance to the pretrained task (\eg, captioning or question answering), it may overfit and generalize poorly to new tasks. Thus, we consider capturing task-agnostic video characteristics as indicators for removing visual redundancy. To this end, 
we leverage the Fast Fourier Transform (FFT) algorithm \cite{cooley1965algorithm} to transform token key vectors to the frequency domain. The core intuition is that token frequency signals content variability -- static or redundant content often shows lower frequency compared to changing and dynamic content. Furthermore, frequency can also be efficiently calculated using FFT.

Specifically, to derive a frequency score that correlates with each token, we first apply FFT along each token dimension across a sequence of token's key representations $\{\mathbf{k}_i\}_{i=1}^p$.
\begin{equation}
    \mathbf{Z}_{\text{fft}} = FFT(\mathbf{k}^{P\times D}), \quad   \mathbf{Z}_{\text{fft}}\in \mathbb{R}^{P\times D}.
\end{equation}
Then, the frequency scores $\mathbf{I}_{\text{fft}}$ are obtained by averaging the magnitudes across all dimensions of each token's frequency representation $\mathbf{Z}_{\text{fft}}$:
\begin{equation}
    \mathbf{I}_{\text{fft}} = Mean (\mathbf{Z}_{\text{fft}}^{P\times D}), \quad   \mathbf{I}_{\text{fft}}\in \mathbb{R}^{P\times 1}.
\end{equation}
We perform simple mean-pooling along dimensions because token of higher frequency often shows larger magnitude (and lower frequency smaller ones) in most dimensions of its frequency representation. 

\textbf{Dual Signal Fusion.} Finally, we fuse the attention and frequency scores via a weighted sum:
\begin{equation}
    \mathbf{I}_{f_t} = \alpha_{f_t} \, \widehat{\mathbf{I}}_{\text{att}} + (1-\alpha_{f_t}) \, \widehat{\mathbf{I}}_{\text{fft}}, \quad  \mathbf{I}_{f_t}\in \mathbb{R}^{P\times 1},
\end{equation}
where $\widehat{\mathbf{I}}$ denotes min-max normalization to $[0,1]$, and $\alpha_{f_t}$ controls the attention importance and frequency score trade-off. Similarly, we can obtain indictor scores 
$\mathbf{I}_{p_{f_t}}\in \mathbb{R}^{\lfloor P/S\rfloor \times 1}$, $\mathbf{I}_{v_t}\in \mathbb{R}^{O\times 1}$
for KV caches at the patch and segment levels, respectively.

% \paragraph
\textbf{Granularity Adaptive Compression.}
According to the fused importance scores $\mathbf{I}_{v_t}$, $\mathbf{I}_{f_t}$, $\mathbf{I}_{p_{f_t}}$, we sort in descending order and keep the top $\kappa_{v_t} = \lfloor \rho_{v_t} \cdot |v_t| \rfloor$, 
$\kappa_{f_t} = \lfloor \rho_{f_t} \cdot |f_t| \rfloor$, $\kappa_{p_{f_t}} = \lfloor \rho_{p_{f_t}} \cdot |p_{f_t}| \rfloor$
tokens, where $\rho_{v_t}$, $\rho_{f_t}$, and $\rho_{p_{f_t}}$ are granularity-specific hyper-parameters. Our motivation is that different granularities serve different functional roles: segment provides semantic narrative contexts, while patch captures regional changes. Thus, we apply different compression ratio  for reducing tokens at different granularities.
Finally, the selected KV slices $\{(\mathbf{K}_p^{(\ell)}, \mathbf{V}_p^{(\ell)})\}_{\ell=1}^{L}$,
$\{(\mathbf{K}_f^{(\ell)}, \mathbf{V}_f^{(\ell)})\}_{\ell=1}^{L}$, $\{(\mathbf{K}_v^{(\ell)}, \mathbf{V}_v^{(\ell)})\}_{\ell=1}^{L}$
of different granularities are stored with their corresponding temporal timestamps, forming a multi-grained yet compact KV cache.

\subsection{Semi-Hierarchical KV-Cache Retrieval}
\label{sec:retrieval}
\begin{figure}[!t]
\centering
\includegraphics[width=0.4\textwidth]{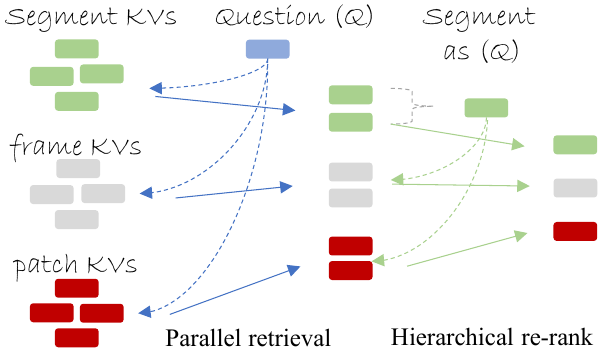}
\caption{Illustration of semi-hierarchical retrieval.}
\label{fig:ret}
\vspace{-3mm}
\end{figure}

For online QA, we utilize a two-stage semi-hierarchical retrieval mechanism to obtain a subset of query-relevant video KV caches. As shown in \cref{fig:ret}, we first retrieve in parallel among KV caches across all three granularities with the question as query, and then rerank the top-ranked KV caches via coarse-to-fine cross-granularity hierarchical retrieval. The final subset of top-ranked KV caches are loaded into LLM for answer decoding. 

Specifically, within each granularity KV cache block (A block can be understood as a frame, a segment, or a patch, depending on what information granularity being processed. We call block as they are incompleted representations because of token compression or pruning), \eg, $\{(\mathbf{K}_{f_t}^{(\ell)}, \mathbf{V}_{f_t}^{(\ell)})\}_{\ell=1}^{L}$, we mean-pool the last-layer key vectors across the block (multi-heads are concatenated), \eg,  
\begin{equation}
    \mathbf{k}_{f_t} = \frac{1}{N_p} \sum_{j=1}^{N_p} \mathbf{k}_j, \quad  \mathbf{k}_{f_t} \in \mathbb{R}^C,
\end{equation}
where $N_p < P$ denotes number of reserved tokens in the frame-level block, and $C=H\times D$ denotes feature dimension. 
Similarly, we can obtain $\mathbf{k}_{v_t}$ and $\mathbf{k}_{p_{f_t}}$.

\textbf{Stage-1: Parallel Retrieval.} 
For parallel retrieval at the first stage, we mean-pool the last-layer question query tokens (multiple heads are concatenated) to obtain the global question representation:
\begin{equation}
    \mathbf{q} = \frac{1}{N_q} \sum_{k=1}^{N_q} \mathbf{q}_k, \quad  \mathbf{q} \in \mathbb{R}^C,
\end{equation}
where $N_q$ is the number of tokens in the question.
At each granularity, we compute the cosine similarity between the question query $\mathbf{q}$ and all video key representations prior to the question-triggered timestamp $t$ (\eg, $\mathbf{K}^{t\times C}_{f_t}$):
\begin{equation}
    \mathbf{s}_{f_t} = \cos\!\left(\mathbf{q}, \mathbf{K}_{f_t}\right), \quad \mathbf{K}_{f_t} \in \mathbb{R}^{t \times C}.
\end{equation}
Then, we retrieve top-$2k_g$ blocks per granularity according to $\mathbf{s}$, where $k_g$ is granularity-specific hyper-parameters.

\textbf{Stage-2: Hierarchical Reranking.} 
The above grain-agnostic parallel retrieval will introduce noises as the model may show undesirably strong cross-model responses on some local visual contents. Thus, to enforce global coherence, we further apply the video representations at high granularity to identity the noises at lower granularities via cross-grained hierarchical retrieval. As shown in \cref{fig:ret} (right),  
to accomplish this, we first obtain a global query by averaging the top-$N$ segment-level representations retrieved at the first stage. 
For block candidate $j$ at lower granularities, we compute a consistency score $\gamma_j$, measuring its alignment with this global query, and update its ranking score via:
\begin{equation}
    \widetilde{s}_j = (1 - \lambda_g) \, s_j + \lambda_g \, \gamma_j,
\end{equation}
where $\lambda_g$ is a coherence enforcement factor. Blocks at lower granularity require higher $\lambda_g$ to anchor them within the global segment, while for segment-level blocks, we set $\lambda_g = 0$ since the global queries are derived from them. Finally, we select the top-$k_g$ blocks per granularity based on this calibrated score $\widetilde{s}$.

\subsection{Question-Answering Using Retrieved KV.} The retrieved video KV caches serve as the context for Video-LLM question-answering. Formally, the attended values are calculated as:
\begin{equation}
    a_t = \text{LLM}(\mathbf{W}_Q\mathbf{X}, [\mathbf{R}_k, \mathbf{W}_K\mathbf{X}], [\mathbf{R}_v, \mathbf{W}_V\mathbf{X}]),
\end{equation}
where $\mathbf{X}$ represents either the question tokens or the current token being decoded, and $\mathbf{R}_k$ and $\mathbf{R}_v$ are the key and value vectors from the context, which include both retrieved video KV and previously generated tokens. The LLM generates answer $a_t$ autoregressively. For subsequent queries, the process repeats with updated timestamp constraints (\eg, $t=t+1$), allowing multiple questions to be answered efficiently in a streaming manner.

%% file: sec/4_experiment.tex
\section{Experiment}
\subsection{Experimental Setup}
\textbf{Datasets and Evaluation.}
We evaluate MuKV on three long streaming video QA datasets: RVSEgo \cite{zhang2025flash}, RVSMovie \cite{zhang2025flash}, and StreamingBench \cite{lin2024streamingbench}. Extended experiments on offline long video QA datasets \cite{fu2025video,zhou2025mlvu,mangalam2023egoschema} are presented in the Supplementary. RVSEgo includes 1.4k questions over 11 ego view videos of averaging 30 minutes. RVS-Movie includes 1.9k questions over 20 movie videos of averaging 1 hour. StreamingBench includes 4.5K questions over 900 videos, in which we focus on the subset which emphasizes real-time visual understanding, including 2.5k questions over 500 videos of averaging 10 minutes. Other dataset details are presented in the Supplementary.

We compare with previous arts under three metrics: online QA accuracy, efficiency,  and offline memory (cache size). For effective accuracy comparison on RVSEgo and RVSMovie, we directly adopt the evaluation script provided by \cite{zhang2025flash,distreaming} which uses LLM as answer judge. Unfortunately, the default LLM judge GPT-3.5-turbo-0613 has been deprecated, so we use GPT-3.5-turbo. This results in slight accuracy differences (specified in \cref{tab:main_res}). On StreamingBench, we report standard multi-choice accuracy. 
For memory and inference efficiency, we report the number of cached or inference visual tokens as a device-independent efficiency metric, \ie, less tokens bring higher efficiency.

\begin{table*}[ht!]
\centering
\small
\setlength{\tabcolsep}{3.5pt}
\caption{Streaming VideoQA performance comparison. For evaluation on VStream-QA \cite{zhang2025flash}, we use GPT-3.5-turbo, and gray out the previous results evaluated using GPT-3.5-turbo-0613 as it was deprecated. Less inference visual tokens and memory tokens indicate higher efficiency. For the number of memory tokens, we report on a per 300-frame (10 minutes) basis, \eg, $59K\approx196*300$.}
\label{tab:main_res}
\vspace{-3mm}
\resizebox{\textwidth}{!}{
\begin{tabular}{lcccccccccccccccc}
\toprule
\multirow{2}{*}{{\bf Model}} & \multirow{2}{*}{{\bf Size}} & \multirow{2}{*}{\#{\bf Inf. Tok} $\downarrow$} & \multirow{2}{*}{\# {\bf Mem. Tok} $\downarrow$} & \multicolumn{2}{c}{{\bf VStream-QA}} & \multicolumn{11}{c}{{\bf StreamingBench (Real-Time)}} \\
\cmidrule(lr){5-6} \cmidrule(lr){7-17}
& & & & {\bf RVSEgo} & {\bf RVSMovie} & {\bf OP} & {\bf CR} & {\bf CS} & {\bf ATP} & {\bf EU} & {\bf TR} & {\bf PR} & {\bf SU} & {\bf ACP} & {\bf CT} & {\bf All} \cr
\midrule
\rowcolor{lightgray}
FVStream \cite{zhang2025flash} & 7B &-- &-- & 59.0 & 56.1 & -- & -- & -- & -- & -- & -- & -- & -- & -- & -- & \cr
\rowcolor{lightgray}
ReKV \cite{distreaming} & 0.5B &12.5K &59K &54.7 & 44.6 & -- & -- & -- & -- & -- & -- & -- & -- & -- & -- & \cr
\rowcolor{lightgray}
ReKV \cite{distreaming} & 7B &12.5K&59K & 63.7 & 54.4 & -- & -- & -- & -- & -- & -- & -- & -- & -- & -- & \cr
\midrule
LongVA (128F) \cite{li2024llava} & 7B & 18.4K & 0  & -- & -- & 70.0 & 63.3 & 61.2 & 70.9 & 62.7 & 59.5 & 61.1 & 53.7 & 54.7 & 34.7 & 60.0 \cr
FVStream \cite{zhang2025flash} & 7B &-- &-- & 58.5 & 54.7 & 27.1 & 45.4 & 25.9 & 25.0 & 28.5 & 13.6 & 19.4 & 26.3 & 24.9 & 50.2 & 24.3 \cr
ReKV \cite{distreaming} & 0.5B &12.5K & 59K & 51.5 & 42.3 & 55.3 & 57.0 & 61.2 & 60.5 & 56.0 & 50.1 & 48.1 & 46.3 & 51.1 & 33.0 & 52.7  \cr
ReKV \cite{distreaming} & 7B &12.5K &59K & 56.2 & 48.2 & 71.3 & 70.3 & 69.1 & 68.6 & 61.6 & 55.2 & 63.9 & 58.1 & 55.7 & 46.3 & 62.3 \cr
\midrule
MuKV (Ours) & 0.5B & 8.3K &59K & 57.9 & 45.2 & 60.7 & 67.2 & 65.0 & 63.5 & 61.0 & 51.7 & 63.0 & 50.1 & 54.0 & 32.4 & 56.8  \cr
MuKV (Ours) & 7B & 8.3K &59K & 59.5 & 48.5 & 74.0 & 78.2 & 72.2 & 71.8 & 64.8 & 55.5 & 66.7 & 58.9 & 59.4 & 39.4 & 64.4  \cr
\bottomrule
\end{tabular}
}
\end{table*}

\textbf{Configurations.}
We utilize the classical backbone model LLaVA-OV \cite{li2024llava} (both 0.5B and 7B) to instantiate our MuKV framework. Our major experiments are conducted on A5000 GPUs with each 24G memory.
We list the values of some key hyper-parameters, which are greedily searched on RVSEgo and applied to the other two datasets. Specifically, for each video, we sample at 0.5FPS unless otherwise indicated, and 4 continuous frames (spanning 8 seconds) constitute a segment, where a frame is divided into $S=4$ super-patches (original patches $P=196$) for multi-grained video representations. 
For retrieval, we retrieve a fixed number of $k_g=64$ blocks following ReKV \cite{distreaming}. Other granularity specific hyper-parameters are:
$\alpha=\{0.5, 0.7, 0.8\}$, $\rho=\{0.1, 0.1, 0.8\}$, $k_g=\{20, 32, 12\}$, and $\lambda_g=\{0.3, 0.3, 0\}$ for patch-, frame-, and segment-level information, respectively.

\subsection{Main Results}
\cref{tab:main_res} compares MuKV with existing long and streaming video QA methods. The results show that MuKV consistently outperforms other recent competitors on RVSEgo and StreamingBench. 
While FVStream \cite{zhang2025flash} wins on RVSMovie, it generalizes extremely poor to StreamingBench. We speculate that FVStream likely overfits the training data since it is additionally learned instead of using existing MLLMs. 
Notably, MuKV achieves stable accuracy improvements over ReKV \cite{distreaming}, without increasing the memory usage and even slightly boosting inference efficiency, demonstrating the high effectiveness and robustness of our innovation designs. In the Supplementary (Sec.~\ref{appsec:offlineQA}), we further show that MuKV's performance can be significantly boosted. For example, the overall accuracy on StreamingBench improves to 71.4\% when we increase the frame rate from 0.5FPS to 3PFS, suggesting its potential for long streaming VideoQA with a large memory budget. Moreover, by replacing the backbone model LLaVA-OV \cite{li2024llava} with Qwen3-VL \cite{bai2025qwen3}, MuKV can gain further improvements on popular long video QA tasks.

On StreamingBench, MuKV achieves remarkable performance improvements over LongVA \cite{zhang2024long} and ReKV \cite{distreaming} on questions about higher-level video understanding, such as ``Causal Reasoning (CR)'', ``Clip Summarize (CS)'' , ``Event Understanding (EU)'', and ``Prospective Reasoning'', but fall short in answering questions of ``Counting (CT)''.  We speculate that our KV cache compression can effectively maintain the global semantics of a video segment but may be ill-suited for counting which is sensitive to fine-grained information changes. 

\begin{figure*}[t!]
    \centering
    \includegraphics[width=1.0\textwidth]{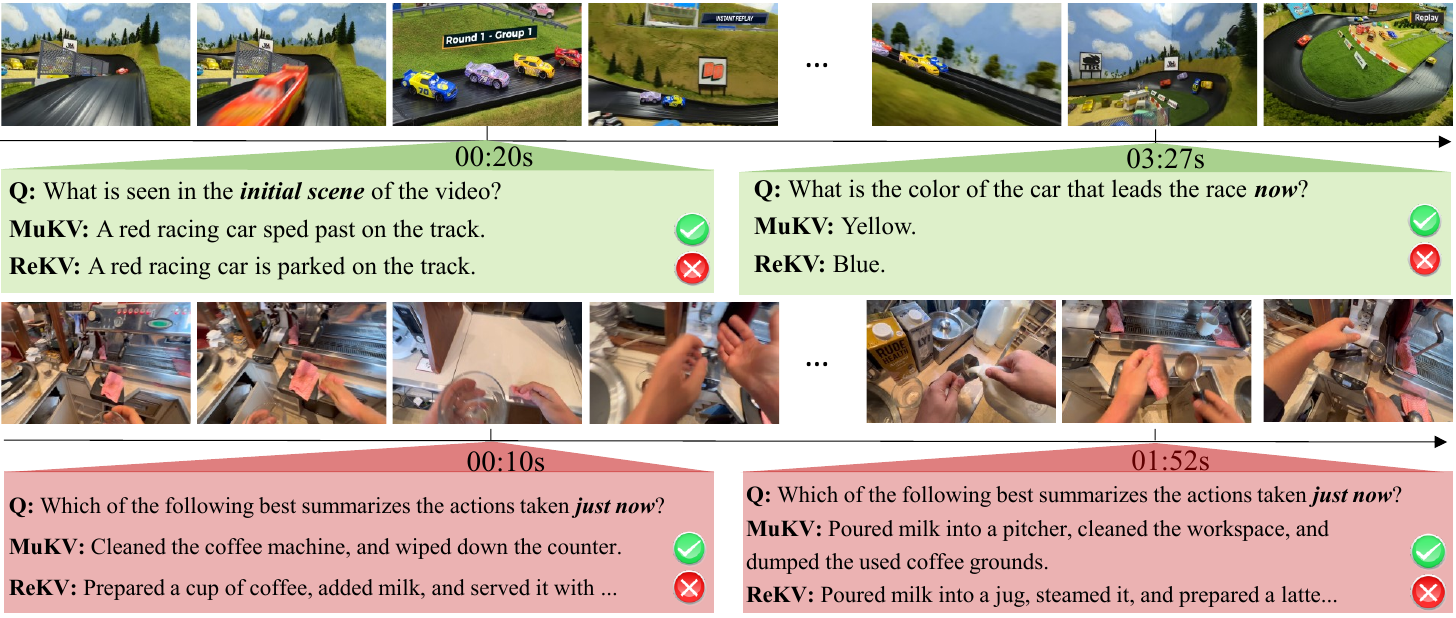}
    \caption{Prediction visualization on StreamingBench \cite{lin2024streamingbench}. Our method (MuKV) outperforms ReKV \cite{distreaming}. It effectively reasons about segment-level temporal dynamics and region level details despite with extended video content inputs.}
    \label{fig:resvis}
    % \vspace{-2mm}
\end{figure*}

Finally, the result visualization in \cref{fig:resvis} demonstrates MuKV's superior behavior in reasoning both segment-level temporal dynamic and region-level details.

\begin{table}[t]
\centering
\small
\setlength{\tabcolsep}{1pt}
\caption{Investigation of MuKV (0.5B) under different granularities. (All variants are under parallel retrieval and are with the same QA efficiency for fair accuracy comparison.)}
\vspace{-3mm}
\resizebox{0.48\textwidth}{!}{%
\begin{tabular}{lccccccc}
\toprule
{\bf Model} & {\bf Pat.} & {\bf Fra.} & {\bf Seg.} &\# {\bf Inf. Tok}$\downarrow$ & \# {\bf Mem. Tok}$\downarrow$ & {\bf Acc@Ego} & {\bf Acc@Movie} \\
\midrule
MuKV & \cmark & \xmark & \xmark & 8.3k & 47k & 51.6 & 44.1  \\
MuKV  & \xmark & \cmark & \xmark & 8.3k & 39k & 53.1 & 45.2 \\
MuKV  & \xmark & \xmark & \cmark & 8.3k & 10k & 54.9 & 44.8 \\
\midrule
MuKV & \cmark & \cmark & \xmark & 8.3k & 53k & 55.1 & 43.6  \\
MuKV & \cmark & \xmark & \cmark & 8.3k & 53k & 55.2 & 45.1 \\
MuKV & \xmark & \cmark & \cmark & 8.3k & 53k & 55.4 & 43.7 \\
\midrule
{\bf MuKV} & \cmark & \cmark & \cmark & 8.3k & 59K & {\bf 56.5} & {\bf 46.0} \\
\bottomrule
\end{tabular}}
\label{tab:granularity}
\vspace{-2mm}
\end{table}
\subsection{Ablation Studies}

\textbf{Multi-granularity.}
\cref{tab:granularity} analyzes the impact of video KV representations at different granularities. Results in the top block show that segment-level KV representations alone generally outperform the other two granularities, suggesting that higher-level and temporally contextual semantics (\eg, actions and events) play a more crucial role in long video understanding. In the middle block, combining segment-level representations with either frame- or patch-level ones consistently yields better results than other combinations and also surpasses those of using a single granularity. Finally, the bottom block shows that integrating all three granularities achieves the best performance, demonstrating the superiority of our multi-grained video KV representations. Notably, our accuracy gains incur no additional memory or online QA overhead owing to the compression mechanism.

\begin{table}[t!]
\centering
\small
\renewcommand{\arraystretch}{0.8}
\setlength{\tabcolsep}{1.5pt}
\caption{Investigation of KV cache compression. Half frame: we adopt a more sparse sampling rate of 0.25 FPS. Rand(50\%): we randomly drop half of tokens. DCP(ratio): we drop $\{$ratio$\}$ of tokens using our DCP compression algorithm which considers both attention and frequency.}
\label{tab:compression_methods}
\vspace{-3mm}
\resizebox{0.48\textwidth}{!}{%
\begin{tabular}{lcccccc}
\toprule
\multirow{2}{*}{{\bf Model}} & \multicolumn{2}{c}{{\bf Compression}} &\multirow{2}{*} {\#{\bf Inf. Tok}$\downarrow$} & \multirow{2}{*}{\#{\bf Mem. Tok}$\downarrow$} & \multirow{2}{*}{{\bf Acc@Ego}} & \multirow{2}{*}{{\bf Acc@Movie}} \\
\cmidrule(lr){2-3} 
 & {\bf Att} & {\bf Freq} &  &  &  &   \cr
\midrule
MuKV &  &   &  12.5K & 177K & 53.7 & 44.3   \\
MuKV &  \cmark &   & 8.3K & 59K & 55.9 & 45.1 \\
MuKV &   & \cmark  & 8.3K & 59K & 56.6 & 45.3 \\
MuKV  & \multicolumn{2}{c}{DCP(67\%)}  & 8.3K & 59K & {\bf57.3} & {\bf45.6} \\
MuKV  & \multicolumn{2}{c}{DCP(50\%)}  & 6.3K & 89K & 53.7 & 43.8 \\
MuKV  & \multicolumn{2}{c}{DCP(75\%)}  & 3.1K & 44K & 54.9 & 45.2 \\
\midrule
ReKV \cite{distreaming} & --& --& 12.5K & 59K & 51.5 & 42.3 \\
ReKV  & \multicolumn{2}{c}{Half Frames} & 12.5K & 29K & 53.0 & 41.9 \\
ReKV & \multicolumn{2}{c}{Rand(50\%)} &12.5K & 29K & 46.7  & 34.7  \\
ReKV & \multicolumn{2}{c}{DCP(25\%)} &9.5K & 44K & 54.3 & 45.2  \\
ReKV & \multicolumn{2}{c}{DCP(50\%)} &6.3K & 29K & {\bf 56.1} & 44.9  \\
ReKV & \multicolumn{2}{c}{DCP(75\%)} &3.1K & 15K & 51.1 & 44.1  \\
ReKV & \multicolumn{2}{c}{DCP(90\%)} &1.3K & 6K & 50.9 & {\bf 46.8}  \\
\bottomrule
\end{tabular}}
\vspace{-2mm}
\end{table}

\begin{table}[t!]
\centering
\small
\renewcommand{\arraystretch}{0.8}
\setlength{\tabcolsep}{1pt}
\caption{Real deployment metrics. s/Q: Average seconds needed for answering each question (online KV retrieval and QA latency). G/h: Cache size per hour of video contents.}
\label{tab:rdm}
\vspace{-3mm}
\resizebox{0.48\textwidth}{!}{%
\begin{tabular}{lccccc}
    \toprule
    {\bf Model} & {\bf Compress Ratio} & {\bf Time (s/Q)$\downarrow$} & {\bf KV-Cache (G/h)$\downarrow$} & {\bf Acc@Ego} \\
    \midrule
    ReKV  & 0 & 0.92 & 4.00 & 51.5 \\
    MuKV  & 0 & 0.72 & 3.72 & 53.7  \\
    {\bf MuKV}  & 2/3 & 0.65 & 1.23 & {\bf 57.3} \\
    MuKV  & 1/2  & 0.67 & 1.85 & 53.7 \\
    MuKV  & 3/4  & 0.59 & 0.91 & 54.9 \\
    \bottomrule
\end{tabular}
}
\end{table}

\textbf{KV Cache Compression.}
We experiment with different models to validate the effectiveness of our DCP compression in \cref{tab:compression_methods}. Results in the top block show that DCP effectively reduces MuKV's memory, and in turn improves QA efficiency and accuracy. Also, both self-attention scores and frequency scores serve as effective indicators for token importance. Notably, when retaining the same KV cache size with ReKV, \eg, compressing the KV cache by 2/3 (67\%), we earn significant performance boost over ReKV, \eg, +5.8\% and +3.3\% on RVSEgo and RVSMovie respectively (Tab.~\ref{tab:rdm} also shows the real deployment performances to better understand this benefit.).  The bottom block shows that DCP consistently improves ReKV \cite{distreaming}. For instance, pruning half of the tokens using DCP (DCP(50\%)) reduces ReKV's memory size by half, and speeds up its online inference by 2$\times$, while also increasing the accuracy by 4.6\% on RVSEgo and 2.6\% on RVSMovie.

To further validate such strength, we randomly drop half of the tokens in ReKV (Rand(50\%)), thus maintaining the same KV cache size. The accuracy, however, decreases by 4.8\%, indicating that our method benefits from informed compression rather than random token reduction. We also test a more sparse frame sampling strategy (Half Frames) to reduce memory usage. While this slightly improves performance on RVSEgo, it leads to a performance drop on RVSMovie. Further dataset analysis (see Supplementary) reveals that the average ratio of answer moments to the video length (before the question timestamp) in RVSEgo is roughly 3$\times$ that of RVSMovie. The smaller answer-moment ratio in RVSMovie makes the task akin to finding a needle in a haystack, where sparse sampling becomes detrimental. Instead, our DCP compression algorithm effectively benefits QA performance on both datasets. 

Additionally, we study other compression ratios and find the optimal one to be 67\% (2/3) for MuKV (We compress the KV cache to be 1/3 of ReKV since we have three granularity representations.). For ReKV, the optimal ratios are 50\% on RVSEgo and 90\% on RVSMovie. The larger compression ratio on RVSMovie indicates its high content redundancy, possibly due to much longer videos, \eg~1 hour.

Finally, we conduct a controlled comparison with the recent video KV-cache compression approach InfinitPot-V \cite{kim2025infinipot}. Tab.~\ref{tab:cexp} shows that our method steadily surpasses both the whole InfinitPot-V approach and its core compression module (TaR and VaN), suggesting DCP's strength for video KV cache compression.
\begin{table}[t!]
\centering
\small
\renewcommand{\arraystretch}{0.8}
\setlength{\tabcolsep}{3.5pt}
\caption{Controlled comparison with existing video KV-cache compression method (\eg, InfinitPot-V \cite{kim2025infinipot}).}
\label{tab:cexp}
\vspace{-3mm}
\begin{tabular}{lcccc}
    \toprule
    {\bf LLaVA-OV-7B} & {\bf \#Inf. Tok$\downarrow$} & {\bf \# Mem. Tok$\downarrow$} & {\bf Acc@Ego} \\
    \midrule
    +ReKV & 12.5K & 5.9K & 55.8 \\
    +InfiniPot-V  & 12.5K & 6K & 57.9 \\
    +MuKV (TaR+VaN) & 8.3K & 5.9K & 58.3 \\
    +{\bf MuKV (DCP)}  & 8.3K & 5.9K & {\bf 59.5} \\
    \bottomrule
\end{tabular}
\end{table}
\begin{table}[t!]
\centering
\small
\setlength{\tabcolsep}{6.1pt}
\renewcommand{\arraystretch}{0.8}
\caption{Comparison between compressing (pruning) tokens of low and high frequency scores.}
\label{tab:fft}
\vspace{-3mm}
\begin{tabular}{lccccc}
\toprule
\multirow{2}{*}{{\bf Model}} & \multicolumn{2}{c}{{\bf Frequency}} &\multirow{2}{*} {{\bf Ratio}}  & \multirow{2}{*}{{\bf Acc@Ego}} & \multirow{2}{*}{{\bf Acc@Movie}} \\
\cmidrule(lr){2-3} 
 & Low & High &  &  &    \cr
\midrule
ReKV  & \xmark & \cmark & 50\% & 55.1 & 43.9 \\
ReKV  & \cmark  & \xmark & 50\% & 51.4 & 41.4 \\
\bottomrule
\end{tabular}
\vspace{-2mm}
\end{table}

\begin{figure}[t!]
    \centering
    \includegraphics[width=0.48\textwidth]{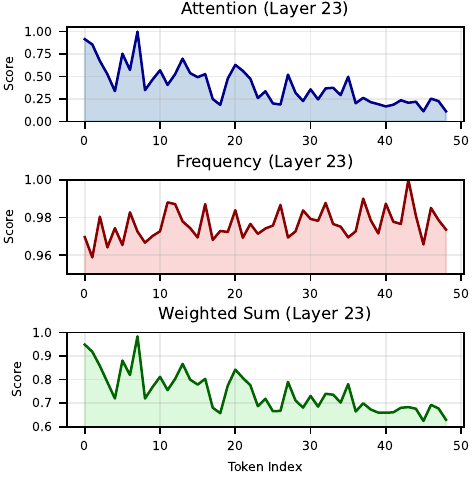}
    \caption{Distribution of tokens' self-attention scores (top), frequency signals (middle), and their weighted sums (bottom).}
    \label{fig:layer23_combined}
    \vspace{-2mm}
\end{figure}

\textbf{Effect of Frequency.} 
We further examine whether retaining high- or low-frequency tokens yields better performance. In videos, high-frequency components typically correspond either to foreground details in static scenes or to rapid motion in dynamic scenes. As shown in \cref{tab:fft}, preserving higher-frequency tokens consistently leads to superior results.
To better illustrate the role of frequency, we visualize the tokens' distributions of self-attention, frequency, and their combined scores from the last LLM layer in \cref{fig:layer23_combined}. Interestingly, earlier coming tokens often receive higher attentions and later tokens lower attentions regardless of samples. In contrast, the frequency distribution is more instance-specific and therefore provides an effective way to calibrate this positional bias of self-attention, ultimately enjoying the best of both and improving compression performance.

\textbf{Semi-hierarchical Retrieval.}
\cref{tab:retr} compares our semi-hierarchical retrieval strategy with both parallel and naive hierarchical retrieval methods. For fair comparison, we retrieve equal number of KV blocks for all three methods. For parallel retrieval, we directly return the top-$k_g$ blocks retrieved at our first stage. For hierarchical retrieval, we retrieve only the frame- and patch-level KV blocks within the top-5 ranked segment blocks, with also the question as query in all stages. The combined KV blocks from all three granularities are loaded to LLM for answer inference. The results show that our semi-hierarchical method achieves the best results on RVSEgo while finds a balance between parallel and hierarchical retrieval on RVSMovie.
\begin{table}[t!]
\centering
\small
\vspace{-3mm}
\setlength{\tabcolsep}{4.5pt}
\caption{Study of different KV retrieval methods.}
\vspace{-3mm}
\begin{tabular}{lccccc}
\toprule
Model & Para. & Hier. & Semi-Hier. & Acc@Ego & Acc@Movie \\
\midrule
MuKV & \cmark &  &   & 56.5 & 46.0 \\
MuKV & & \cmark & & 52.9 & 43.1 \\
MuKV & & & \cmark & 57.9 & 45.2 \\
\bottomrule
\end{tabular}
\label{tab:retr}
\vspace{-2mm}
\end{table}

\textbf{LLM Layer Investigation.}
While we derive the attention and frequency scores from the last LLM layer (so as for KV cache retrieval), we also investigate other alternatives.
Tab.~\ref{tab:layerab} shows that all other designs perform worse than the last layer representations. This is reasonable as the last-layer representations are closer to our target task that our backbone MLLMs are designed for.
\begin{table}[t!]
\centering
\small
\vspace{-3mm}
\setlength{\tabcolsep}{2pt}
\caption{Layer investigation on RVS-Ego.}
\label{tab:layerab}
\vspace{-3mm}
\resizebox{\linewidth}{!}{%
\begin{tabular}{l|cccc}
\toprule
{\bf LLaVA-OV-0.5B} & {\bf Last (default)} & {\bf Penultimate} & {\bf Middle} &  {\bf Last-5 Fusion} \\
\midrule
+MuKV & 57.3  & 54.9 & 53.4 & 56.5\\
\bottomrule
\end{tabular}
}
\vspace{-3mm}
\end{table}

%% file: sec/5_conclusion.tex
\section{Conclusion}
We advance long streaming VideoQA by introducing MuKV, a framework that compactly stores and effectively retrieves multi-grained video KV caches for accurate and efficient answer decoding. To enable compact storage offline, we propose a dual-signal KV-cache compression module that jointly considers token self-attention importance and token frequency in the Fourier domain. For effective online retrieval, we design a semi-hierarchical retrieval mechanism that performs grain-agnostic parallel retrieval followed by cross-grain hierarchical re-ranking. Extensive experiments demonstrate that MuKV significantly improves accuracy without increasing offline memory or online inference latency, revealing substantial redundancy in existing streaming QA methods and highlighting our effectiveness for elimination towards enhancements.

\section*{Acknowledgements}
This research is supported by the Ministry of Education, Singapore, under its MOE Academic Research Fund Tier 2 (MOE-T2EP20125-0037).

%% file: sec/X_suppl.tex
\clearpage
\setcounter{page}{1}
\maketitlesupplementary

\section{Dataset Introduction}
\begin{figure}
  \centering
  \begin{subfigure}{\linewidth}
    \includegraphics[width=\textwidth]{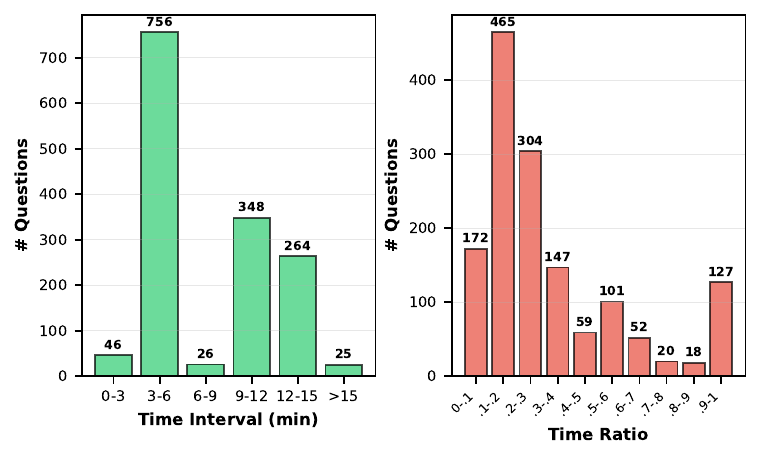}
    \caption{Distribution of answer spans and relative ratios on RVS-Ego.}
    \label{fig:appshort-ego}
  \end{subfigure}
  \hfill
  \begin{subfigure}{\linewidth}
    \includegraphics[width=\textwidth]{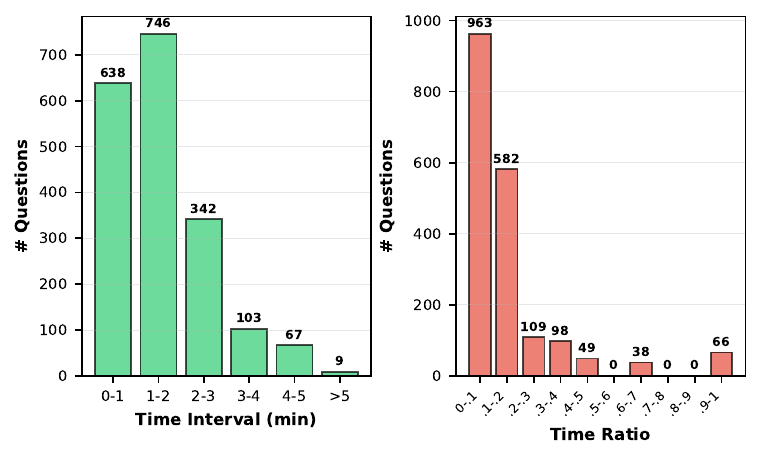}
    \caption{Distribution of answer spans and relative ratios on RVS-Movie.}
    \label{fig:appshort-mov}
  \end{subfigure}
  \caption{The answer spans and their ratios relative to the corresponding question timestamps in RVS-Ego are substantially longer than those in RVS-Movie, meaning that RVS-Ego has a higher chance of capturing the answer segment and thus brings less redundancy under uniform video sampling for streaming QA.}
  \label{fig:appdataset}
  \vspace{-3mm}
\end{figure}

VStream-QA \cite{zhang2025flash} comprises two long-video datasets: RVS-Ego and RVS-Movie. \textbf{RVS-Ego} contains 10 egocentric videos with an average duration of 30 minutes, while \textbf{RVS-Movie} includes 22 movie videos averaging 1 hour. The distributions of the temporal answer spans and their ratios relative to the question timestamps of both datasets are presented in \cref{fig:appdataset}. The results show that the answer spans and their relative ratios in RVS-Ego are substantially longer than those in RVS-Movie, indicating that RVS-Ego has a higher chance of capturing the answer segment and would result in less redundancy under uniform video sampling for streaming QA.
\textbf{StreamingBench} \cite{lin2024streamingbench} does not provide answer span annotations, so we introduce the 10 question categories defined over the subset of 500 videos (on average 10 minutes) in the task of real-time visual understanding:
\begin{itemize}
    \item \textbf{Object Perception (OP):} Detect and identify specific objects, \eg, ``What is the person holding right now?''.
    \item \textbf{Causal Reasoning (CR):} Analyze event cause-and-effect relationships, \eg, ``Why Mr Bean is shocked now?''.
    \item \textbf{Clips Summarization (CS):} Summarize main content in specific video clips, \eg, ``Which of following best summarize the actions just now''.
    \item \textbf{Attribute Perception (ATP):} Identify and categorize object or individual attributes, \eg, ``What color is the car directly in front right now?''
    \item \textbf{Event Understanding (EU):} Recognize and describe sequences of events, \eg, ``What is happening in the initial scene of the video?''
    \item \textbf{Text-Rich Understanding (TR):} Interpret and explain text-rich content within the video, \eg, ``Which team is leading in the racing points?''.
    \item \textbf{Prospective Reasoning (PR):} Predict future events based on current video context, \eg, ``What might the speaker explain next?''.
    \item \textbf{Spatial Understanding (SU):} Understand and describe spatial relationships and locations, ``Where is ... now?''
    \item \textbf{Action Perception (ACP):} Identify specific actions in the video, \eg, ``What is the person doing now?''.
    \item \textbf{Counting (CT):} Count occurrences of specific objects or actions, \eg, ``How many times does .. so far?'' 
\end{itemize}
 
%------------------------------------------------------------------------
\section{Experiments}

\subsection{Offline VideoQA and Different Backbones}
\label{appsec:offlineQA}
We also extend our method MuKV to the popular offline long VideoQA datasets: Video-MME \cite{fu2025video}, MLVU \cite{zhou2025mlvu} and EgoSchema \cite{mangalam2023egoschema}. For a streaming QA setting, we assume that all questions are asked at the end of the videos. The results in \cref{tab:offlineqa} show that MuKV steadily improves over ReKV and other recent streaming QA methods under different backbones, demonstrating the robustness of our multi-grained KV cache compression and semi-hierarchical retrieval approach.
\begin{table*}[t]
  \centering
  \small
  \caption{Results comparison on offline long VideoQA benchmarks. Compared results are copied from StreamMem \cite{yang2025streammem}. For the number of memory tokens, we report on a per 30-frame (1 minute) basis to satisfy all dataset, \eg, $5.9K\approx196*30$.}
  \begin{tabular}{lccccccc}
    \toprule
    \textbf{Method} & \textbf{Frames/FPS} & \textbf{\# Mem. Tok} & \textbf{MLVU} & \textbf{EgoSchema} & \multicolumn{3}{c}{\textbf{VideoMME}} \\
    \cmidrule(lr){6-8}
     & & & & & \textbf{Medium} & \textbf{Long} & \textbf{All} \\
    \midrule
    GPT-4o \cite{hurst2024gpt} & -- & -- & 64.6 & 72.2 & 70.3 & 65.3 & 71.9 \\
    \midrule
    MovieChat+ \cite{song2025moviechat+} & 2048 & - & 25.8 & 53.5 & - & 33.4 & 38.2 \\
    Dispider \cite{qian2025dispider} & 1 fps & - & 61.7 & 55.6 & 53.7 & 49.7 & 57.2 \\
    \midrule
    LongVA \cite{zhang2024long} & 128 & -- & -- & --  & 50.4 & 46.2 & 52.6 \\
    LongVU \cite{shen2024longvu} & 400/1fps & -- & 67.6 & 58.2 & 59.5 & 60.6 & - \\
    \midrule
    LLaVA-OV-7B \cite{li2024llava} & 32 & -- & 64.7 & 60.1 & 54.7 & 46.2 & 56.9 \\
    \quad + ReKV &  0.5 fps & 5.9K & 68.5 & 60.7 & -- & -- & --  \\
    \quad + LiveVLM \cite{ning2025livevlm} & 0.5/0.2 fps & - & 66.3 & 63.0 & 56.4 & 48.8 & 57.3 \\
    \quad + StreamMem \cite{yang2025streammem} & 0.5/0.2 fps & 6K & 66.9 & 63.0 & 56.6 & 50.1 & 59.4 \\
    \rowcolor{gray!10} \quad + MuKV (Ours) & 0.5fps & 5.9K & 67.8 & 63.3 & 57.9 & 52.1 & 61.2 \\
    \midrule
    LLaVA-OV-0.5B \cite{li2024llava} & 32 & -- & 50.4 & 26.8 & 39.7 & 46.2 & 45.4 \\
    \quad +  ReKV & 0.5 fps & 5.9K & 53.2 & 29.6 & 40.1 & 47.9 & 48.2  \\
    \rowcolor{gray!10} \quad + MuKV (Ours) & 0.5fps & 5.9K & 55.2 & 30.5  & 41.4  & 48.5 & 49.1 \\
    \midrule
    Qwen2.5-VL-3B \cite{bai2025qwen2} & 768 & -- & 63.3 & 64.4 & 58.0 & 47.2 & 60.3 \\
    \quad + InfiniPot-V \cite{kim2025infinipot} & 768 & 6K & 62.1 & 61.8 & - & - & 59.3\\
    \quad + StreamMem \cite{yang2025streammem} & 4.0/0.5 fps & 6K & 62.3 & 62.2 & 60.1 & 49.1 & 59.5\\
    \rowcolor{gray!10} \quad + MuKV (Ours) & 0.5 fps & 5.9K & 63.0 & 63.0 & 61.0 & 50.0 & 60.8\\
    \midrule
     Qwen3-VL-4B \cite{yang2025qwen3} & 768 & -- & 64.8 & 65.8 & 60.2 & 49.5 & 62.5 \\
    \rowcolor{gray!10}\quad + MuKV & 0.5fps & 5.9K & 66.0 & 67.0 & 61.8 & 51.0 & 63.6\\
    \midrule
     Qwen3-VL-2B \cite{yang2025qwen3} & 768 &-- & 64.0 & 65.0 & 59.0 & 48.5 & 61.5 \\
    \rowcolor{gray!10}\quad + MuKV & 0.5fps & 5.9K & 65.2 & 66.3 & 60.8 & 49.8 & 62.6 \\
    \bottomrule
  \end{tabular}
  \label{tab:offlineqa}
\end{table*}

\subsection{Hyper-parameters}

\paragraph{Frame Sampling Rates.} 
\cref{tab:fps_res} shows that denser video sampling improves model performance on StreamingBench but harms performance on RVS-Ego. Upon examining the datasets, we find that RVS-Ego exhibits much lower visual appearance variation than StreamingBench, meaning that sparse sampling already captures most key frames, while denser sampling introduces unnecessary redundancy.
Meanwhile, MuKV consistently improves ReKV~\cite{distreaming} across all sampling rates, with the performance gains becoming more pronounced at higher FPS values. This further demonstrates the effectiveness of our KV compression mechanism for removing redundancy.
\begin{table}[t]
\centering
\small
\setlength{\tabcolsep}{6pt}
\caption{Sensitivity analysis on granularity-specific KV retention (reversed side of compression) ratios $(\rho_p,\rho_f,\rho_s)$. 
Higher segment retention ($\rho_s$) leads to better performance, highlighting the importance of segment-level signal cues.}
\begin{tabular}{ccc|cc|cc}
\toprule
\multicolumn{3}{c|}{Retention Ratio $(\rho_p,\rho_f,\rho_s)$} &
\multicolumn{2}{c|}{RVSEgo} &
\multicolumn{2}{c}{RVSMovie} \\
$\rho_p$ & $\rho_f$ & $\rho_s$ & Acc & Score & Acc & Score \\
\midrule
0.1 & 0.1 & 0.8 & \textbf{57.9} & \textbf{3.89} & {\bf45.2} & {\bf3.34} \\
0.1 & 0.8 & 0.1 & 55.6 & 3.80 & 44.1 & 3.33 \\
0.8 & 0.1 & 0.1 & 54.7 & 3.79 & 43.5 & 3.30 \\
\bottomrule
\end{tabular}
\label{tab:gra_ratio_dcp}
\end{table}

\begin{table}[t]
\centering
\small
\setlength{\tabcolsep}{5pt}
\caption{Sensitivity analysis of $\lambda$ in the semi-hierarchical retrieval module. Smaller $\lambda$ reduces dependency on the second-stage cross-grain retrieval scores, which slightly improves performance on RVSEgo but declines performance on RVS-Movie.}
\begin{tabular}{lcc|cc}
\toprule
\multirow{2}{*}{Method} & 
\multicolumn{2}{c|}{RVSEgo} & 
\multicolumn{2}{c}{RVSMovie} \\
& Acc & Score & Acc & Score \\
\midrule
MuKV rerank ($\lambda$=\{0.7, 0.7, 0\}) & 56.1 & 3.86 & {\bf47.2} & {\bf3.41} \\
MuKV rerank ($\lambda$=\{0.3, 0.3, 0\}) & \textbf{57.9} & \textbf{3.89} & 45.2 & 3.34 \\
\bottomrule
\end{tabular}
\label{tab:gra_lambda_ret}
\end{table}

% We compare with previous methods at different frame sampling rates. 
\begin{table*}[ht!]
\centering
\small
\setlength{\tabcolsep}{3.5pt}
\caption{Streaming VideoQA performance comparison under different video sampling rates (FPS). For evaluation on VStream-QA \cite{zhang2025flash}, we use GPT-3.5-turbo. Less inference visual tokens and memory tokens indicate higher efficiency. For the number of memory tokens, we report on a per 300-frame (10 minutes) basis to satisfy all datasets, \eg, $59K\approx196*300$.}
\label{tab:fps_res}
\resizebox{\textwidth}{!}{
\begin{tabular}{lcccccccccccccccc}
\toprule
\multirow{2}{*}{\textbf{Model}} & \multirow{2}{*}{\textbf{Size}} & \multirow{2}{*}{\textbf{FPS}} & \multirow{2}{*}{\textbf{\#Inf. Tok} $\downarrow$} & \multirow{2}{*}{\textbf{\#Mem. Tok} $\downarrow$} & \multirow{2}{*}{\textbf{RVS-Ego}} & \multicolumn{11}{c}{\textbf{StreamingBench (Real-Time Understanding)}} \\
\cmidrule(lr){7-17}
& & & & & & OP & CR & CS & ATP & EU & TR & PR & SU & ACP & CT & All \cr
\midrule
% ReKV \cite{distreaming} & 0.5B & 0.5 & 12.5K & 59K & 51.5  & 55.3 & 57.0 & 61.2 & 60.5 & 56.0 & 50.1 & 48.1 & 46.3 & 51.1 & 33.0 & 52.7  \cr
% MuKV (Ours) & 0.5B &0.5 & 8.3K &59K & 57.9 & 60.7 & 67.2 & 65.0 & 63.5 & 61.0 & 51.7 & 63.0 & 50.1 & 54.0 & 32.4 & 56.8  \cr
% \midrule
ReKV \cite{distreaming} & 7B & 0.5 & 12.5K &59K & 56.2  & 71.3 & 70.3 & 69.1 & 68.6 & 61.6 & 55.2 & 63.9 & 58.1 & 55.7 & 46.3 & 62.3 \cr
ReKV \cite{distreaming} & 7B & 2 & 12.5K &236K & 54.8 & 72.0 & 70.9 & 66.2 & 69.2 & 63.0 & 55.7 & 64.1 & 58.7 & 56.2 & 46.4 & 62.9 \cr
ReKV \cite{distreaming} & 7B & 3 & 12.5K &354K & 53.9  & 71.0 & 70.5 & 65.2 & 69.2 & 62.1 & 55.2 & 64.5 & 58.2 & 55.8 & 46.6 & 62.5 \cr
MuKV (Ours) & 7B & 0.5 & 8.3K &59K & 59.5 & 74.0 & 78.2 & 72.2 & 71.8 & 64.8 & 55.5 & 66.7 & 58.9 & 59.4 & 39.4 & 64.4  \cr
MuKV (Ours) & 7B & 2 & 8.3K &118K & 57.7  & 78.4 & 78.3 & 82.1 & 77.7 & 68.3 & 57.4 & 70.1 & 62.8 & 61.3 & 42.0 & 68.2  \cr
MuKV (Ours) & 7B & 3 & 8.3K &354K & 56.1 & 82.1 & 82.8 & 82.1 & 80.0 & 71.9 & 61.6 & 74.0 & 65.4 & 65.9 & 43.7 & 71.4  \cr
\bottomrule
\end{tabular}
}
\end{table*}

\paragraph{Granularity Hyper-Parameters.}
\cref{tab:gra_ratio_dcp} studies different compression ratios at KV cache of different granularities. The results show that pruning more KVs at the lower-level granularity often yields better performance, highlighting the strategy of segment-level modeling in video understanding.
\cref{tab:gra_lambda_ret} studies the trade-off parameter between the first-stage parallel-retrieval score and the second-stage cross-grain hierarchical retrieval score. A smaller $\lambda$ brings better performance on RVS-Ego but worse performance on RVS-Movie, suggesting that cross-grain retrieval scores are less effective on egocentric videos which have less event-level content variations compared to that on movie videos.

% . In contrast, a larger $\lambda$ brings better performance on RVS-Movie, indicating that 

%-------------------------------------------------------------------------